\documentclass[conf]{new-aiaa}
\usepackage[utf8]{inputenc}
\usepackage{float}

\usepackage{graphicx}
\usepackage{subcaption}

\usepackage{amsmath}

\usepackage{algorithm}
\usepackage{algpseudocode}
\algnewcommand{\algorithmicand}{\textbf{ and }}
\algnewcommand{\algorithmicor}{\textbf{ or }}
\algnewcommand{\OR}{\algorithmicor}
\algnewcommand{\AND}{\algorithmicand}
\algnewcommand{\var}{\texttt}

\usepackage{longtable,tabularx}

\graphicspath{ {figures/} }

\usepackage[version=4]{mhchem}
\usepackage{siunitx}

\algnewcommand{\leftcomment}[1]{$\triangleright$ #1} 

\makeatletter
\newenvironment{breakablealgorithm}
  {
   \begin{center}
     \refstepcounter{algorithm}
     \hrule height.8pt depth0pt \kern2pt
     \renewcommand{\caption}[2][\relax]{
       {\raggedright\textbf{\ALG@name~\thealgorithm} ##2\par}
       \ifx\relax##1\relax
         \addcontentsline{loa}{algorithm}{\protect\numberline{\thealgorithm}##2}%
       \else
         \addcontentsline{loa}{algorithm}{\protect\numberline{\thealgorithm}##1}%
       \fi
       \kern2pt\hrule\kern2pt
     }
  }{
     \kern2pt\hrule\relax
   \end{center}
  }
\makeatother
\title{GTOC X: Karmarkar's Gang's Approach and Results}

\author{
Aniket Bhushan\footnote{Siemens Digital Industries, \href{mailto:aniket2701bhushan@gmail.com}{aniket2701bhushan@gmail.com}},  
Vishesh Vatsal\footnote{Independent, \href{mailto:vishesh.vatsal20@gmail.com}{vishesh.vatsal20@gmail.com}}, 
Sri Anish Vutukuri\footnote{IISC, \href{mailto:srianish0705@gmail.com}{srianish0705@gmail.com}}, 
C Barath\footnote{Independent, \href{mailto:barathclm@gmail.com}{barathclm@gmail.com}}, 
Abhijit Bannerji\footnote{IISC, \href{mailto:AVIJITINSTRU@gmail.com}{AVIJITINSTRU@gmail.com}}}

\begin{document}

\maketitle

\begin{abstract}
This paper describes the methods used and the results obtained by team Karmarkar's Gang for the 10th edition of the Global Trajectory Optimization Competition. The methods used by our team are described. These methods involve- mothership targeting flyby to target high-value stars using a single impulse, fast ships targeting the edge of the galaxy and settler ships having a fast expansion, time-optimal three impulse transfer strategy to select the targets to get fast and wide spatial distribution. It is seen that there is a scope of improvement with respect to multi-star targeting for motherships. The results of the strategy are discussed.
\end{abstract}

\section{Introduction}
\vspace{1em}

This paper provides one of the possible solutions for GTOC X. GTOC (Global Trajectory Optimization Competition) is an event with world wide challenge \cite{petropoulosgtoc} to solve a “nearly-impossible” problem of interplanetary trajectory design. The winners of the event get the formulate the problem for the next event. The 10th edition of the problem was formulated by Mission Design and Navigation Section, JPL, Caltech. The problem statement was to settle as many stars as possible, in a uniform distribution as possible, while using as little propulsive velocity change as possible. The settlement of the galaxy is to be initiated by the three Mother ships, two Fast ships each with its limitations on magnitude and number of $\Delta$V. Each Mothership is allowed three impulsive maneuvers, each impulse less than 200 km/s, the cumulative impulse less than 500 km/s, and at least 1 Myr of time between two impulses. Either Mothership flybys a target star and ejects a Settlement Pod allowed 1 impulse of up to 300 km/s to rendezvous with a star. Fast Ships are allowed two impulses with a cumulative impulse of less than 1500 km/s. Our approach is to send the three mother ships in angles as separated from each other as possible to provide angular uniformity and the fast ships to outer edges of the galaxy to provide more radial uniformity. With each settled star, up to three Settler ships can depart from each settled star at least 2 Myr after settling and are allowed 5 impulses, each up to 175 km/s and cumulative impulse up to 400 km/s. Our strategy for Settler ships is to start occupying nearby available stars as soon as possible. Each settler ship transfer is optimized for a three impulse trajectory. 

\section{Exploratory Data Analysis}
\vspace{1em}

\begin{figure}[h!]
    \centering
    \begin{subfigure}[b]{0.48\linewidth}
        \includegraphics[width=\linewidth]{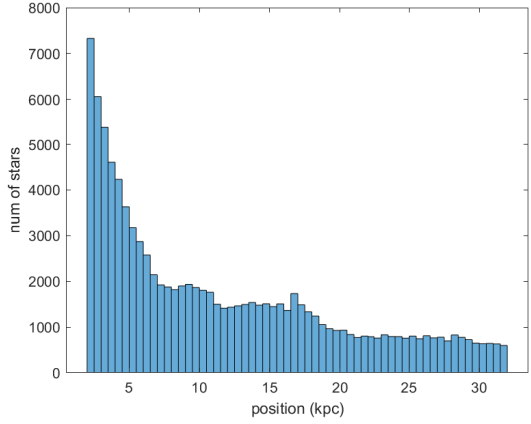}
        \caption{Star distribution versus radial position.}
    \end{subfigure}
    \begin{subfigure}[b]{0.48\linewidth}
        \includegraphics[width=\linewidth]{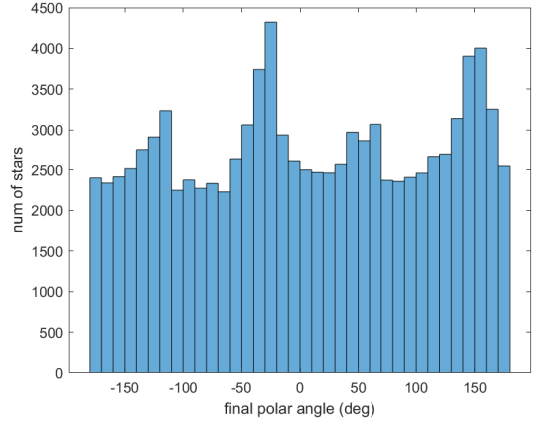}
        \caption{Star distribution versus angular position at t = $90Myr$}
    \end{subfigure}
    \caption{Star distribution versus position}
    \label{fig:starDistributionVsPositionFig}
\end{figure}

The data set contains 100,000 stars in a circular orbit about the center of the galaxy. A histogram of the stars at various positions (between 2 and 32 kpc) is shown in Figure \ref{fig:starDistributionVsPositionFig}. This indicates that star densities go progressively lower versus the position but there is no such trend for stars versus the polar angle. This provides an idea that settling stars farther away will be more difficult due to low star densities.

\newpage

\begin{figure}[h]
    \centering
    \includegraphics[width=0.8\linewidth]{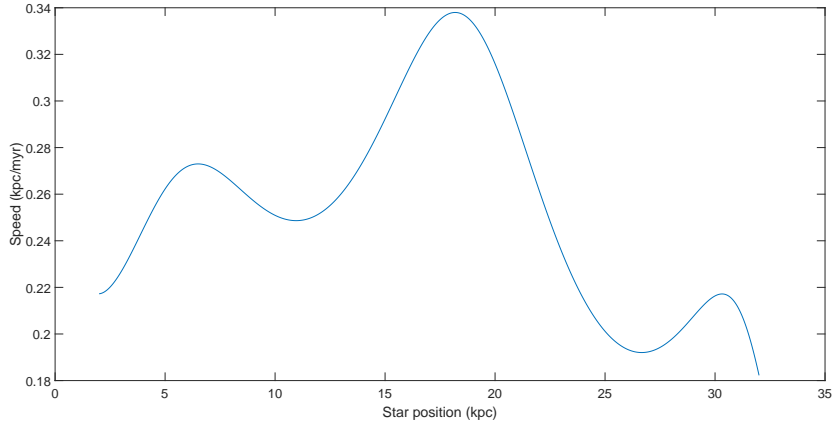}
    \caption{Star speed versus position}
    \label{fig:speedVsStarDistributionFig}
\end{figure}

A quick look at the speed distribution suggests that the stars between 18 and 20 kpc have the maximum speed. There are two other speed maximums seen in the data as well. The mother ships begin from Sol which is around 8 kpc which is near a speed minimum. 

\subsection{Merit function analysis}
\vspace{1em}

The merit function for the problem is defined as:

\begin{equation} \label{EqMeritFn}
\hspace{2cm}
    J=\dfrac{N}{1+10^{-4}N(E_r+E_\theta)} \dfrac{\Delta V_{max}}{\Delta V_{used}} 
\end{equation}

where,
{\renewcommand\arraystretch{1.0}
\begin{longtable*}[ht]{l @{\quad=\quad} l}
    N & Number of settled stars \\
    $\Delta V_{max}$ & Maximum permitted Delta V \\
    $\Delta V_{used}$ & Utilized Delta V for settling N star \\
\end{longtable*}}

Error functions $E_r$ and $E_{\theta}$ are as defined in the problem statement.Broadly these error functions indicate asymmetry of settling in the radial and polar dimensions respectively. 

The following analysis indicates value of the merit function (without 
$\Delta V$ factor) if the first 'N' stars closest to the galactic center start getting populated. We can see that as expected, initially the more the number of stars, the better the merit function. However, it starts to decrease beyond 80,000 stars which suggests that the error functions begin to dominate. 

\vspace{-0.5cm}
\begin{figure}[h!]
    \centering
    \includegraphics[width=0.6\linewidth]{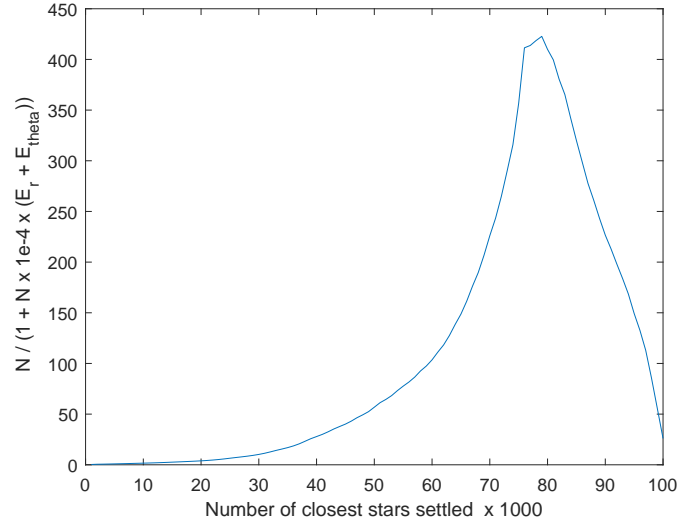}
    \caption{Merit function versus number of closest stars settled}
    \label{fig:meritFnVsNumSettledStarFig}
\end{figure}

\subsection{Identification of high value stars}
\vspace{1em}

To reduce the number of stars to be settled in a limited time of 90 MYrs, it was decided to identify the set of high-value stars. The high-value stars seem to be concentrated at the center of $r$,$\theta$ grids as per our analysis of the radial and polar error functions. Over this, it is required that for transfers to be consuming low $\Delta V$, the target stars should be close to the plane. Hence the selection of the high-value stars is done by identifying the stars closest to the centers of their $r$,$\theta$ grid while having an inclination less than 10 degrees. The following target star database was generated.

\begin{figure}[h!]
    \centering
    \includegraphics[width=0.9\linewidth]{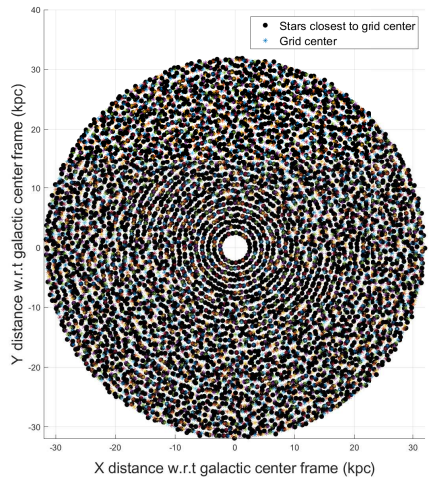}
    \caption{Target stars (marked in black) w.r.t. grid centers identified in color }
    \label{fig:targetStarsWrtGridCentersFig}
    \vspace{-0.5cm}
    \end{figure}
\vspace{-0.5cm}

\section{Formulation}
\vspace{1em}

The star position data was broken into steps of 0.5 MYr, hence providing position of the full 100,000 star database at t=0, 0.5, 1, .... 90 Myrs. Individual strategies for the fast ships, mother ships and settler ships were chosen. 

The shooting solver, settler ship strategy need to be detailed.

\section{Fast Ship strategy}
\vspace{1em}

There are two fast ships available, which has capabilities to give maximum $\Delta V$s as compared to other types of ships. Our strategy is to use these high $\Delta V$ values to reach the edges of the galaxy which is impossible to reach using mother ships.

We define following control parameters for fast ships : 
{\renewcommand\arraystretch{1.0}
\noindent\begin{longtable*}{@{}l @{\quad=\quad} l@{}}
$t_{departure, i}$ & Time at which ith fast ship will depart from sol, (time for first impulse) \\
$[r^{min}_{i}, r^{max}_{i}]$ and $[\theta^{min}_{i}, \theta^{max}_{i}]$ & Identifies the truncated sector that defines the search space for $i^{th}$ FS.
\end{longtable*}}

\begin{figure}[h!]
\centering
\includegraphics[height=5cm,width=6cm]{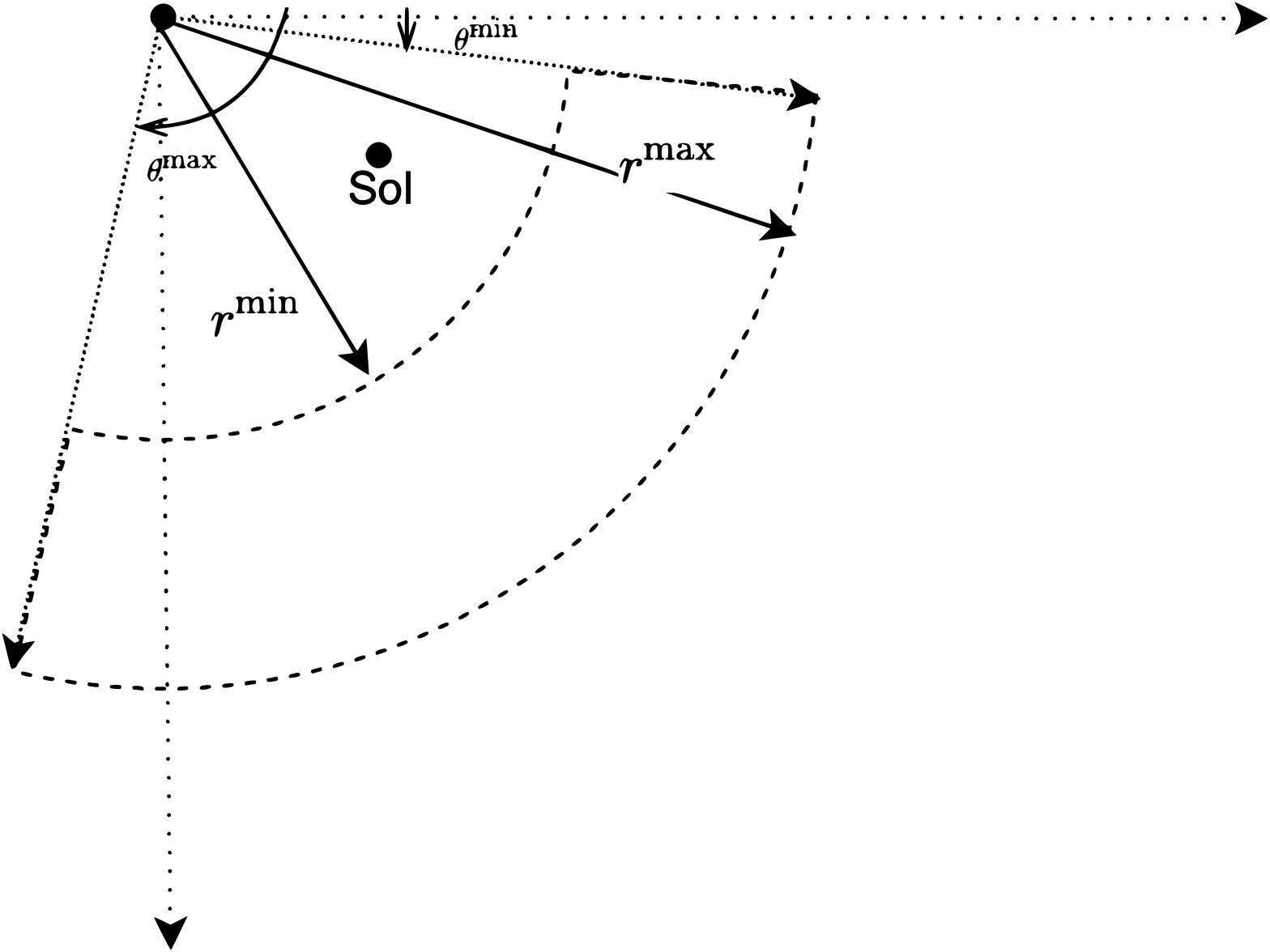} 
\caption{Controlled region to select target stars}
\label{fig:FsStrategyFig}
\end{figure}

\begin{table}[ht]
\caption{Control parameter values for Fast ships}
\centering
\begin{tabular}{c c c c c c}
\hline\hline
FS id & t\textsubscript{departure} & $r^{min}_{i}$ & $r^{max}_{i}$ &  $\theta^{min}_{i}$ &  $\theta^{max}_{i}$ \\
& (Myr) & (Kpc) & (Kpc) & (deg) & (deg) \\[0.5ex]
\hline
1 & 0 & 27 & 27.1 & -180  & -90  \\
2 & 0 & 27 & 27.1 & -90   &  0   \\ [1ex]
\hline
\end{tabular}
\label{table:fastShipControlParamTable}
\end{table}

Once, we have identified a start from the region of interest which satisfies all the constraints, we add these stars to the settlement tree. The fast ship selection strategy allows fastest transfer to stars near the edge and we consider this triggers a fast expansion from the edges towards the inner regions.

\begin{breakablealgorithm}
  \caption{Fast ship Star selection strategy}\label{alg:fsstrategyFn}
  \begin{algorithmic}[1]
    \Procedure{FastShipTransferStrategy}{$r,\theta, t_{departure}$}\Comment{Star selection procedure for fast ship}
      \State $t_{arrival} \gets t_{departure}$ + 0.5 Myr
      \State starPos $\gets$  Position of all stars at $t_{arrival}$
      \State filteredStars $\gets$  Stars - within  $[r^{min}, r^{max}]$ at $t_{arrival}$
      \State
      
      \For {iStar in filteredStars}
        \State $t_{departure}$ $\gets$ 0
        \State $t_{arrival}$ $\gets$ $t_{departure}$ + 2.5 Myr
        \State $tof \gets t_{arrival} - t_{departure}$ 
        \State $r_0, v_{0, guess} \gets$ position and velocity of sol at $t_{departure}$
        \State
        
         \While{true}
           \State $r_t, v_t \gets $ Position \& Velocity of iStar at $t_{arrival}$ 
           \State $\theta_t \gets$ Polar angle of iStart at $t_{arrival}$
           \State 
           
           \If {$\theta_t \in [\theta^{min}, \theta^{max}]$ }
             \State [state, $v_0, v_f$] $\gets$ ShootingSolver($r_0, r_t, tof, v_{0, guess}$ )
             \If {Impulse constraints not satisfied for FS}
               \State $tof += 2.5 Myr$
             \Else
               \State Add to set of reachable stars
               \State break;
             \EndIf
           \EndIf
         \EndWhile
       \EndFor
       \State Sort the stars based on tof
       \State Get the first star
       \State \textbf{return} [$id, \Delta V_1, \Delta V_2, t_{arrival}$]
    \EndProcedure
  \end{algorithmic}
\end{breakablealgorithm}

\section{Mother ship strategy}
\vspace{1em}

We define control parameters for each of the mother ships. These parameters are responsible for defining at what point of time after $t_0$, mother ships will depart from sol, their direction, magnitude of impulse and timings for successive impulses. These control parameters are chosen in such a way that the mother ships can cover as much angular distribution as possible.

{\renewcommand\arraystretch{1.0}
\begin{longtable*}{@{}l @{\quad=\quad} l@{}}
    $t_{departure, j}$     & Time at which mother ship will depart from sol, (time for first impulse) \\
    $t_{ij}$               & Coasting time after i\textsubscript{th} impulse for mother ship j \\
    $\Delta V_{ij}$        & Magnitude of i\textsubscript{th} impulse for mother ship j \\
    $\Delta \theta _{ij} $ & Direction of i\textsubscript{th} impulse from velocity for mother ship j, measured in ACW sense\\
\end{longtable*}}

\begin{table}[ht]
\caption{Control parameter values for Mother ships}
\centering
\begin{tabular}{c c c c c c c c c c c }
\hline\hline
MS & $t_{departure}$ & $t_1$  & $t_2$ & $t_3$ & $\Delta V_{1}$  & $\Delta V_{2}$ & $\Delta V_{3}$ & $\theta _{1}$ & $\theta _{2}$ & $\theta _{3}$ \\
id & (Myr) & (Myr) & (Myr) & (Myr) & (km/s) & (km/s) & (km/s) & (deg) & (deg) & (deg) \\[0.5ex]
\hline
1 & 0 & 10 & 5  & 15 & 100 & 100 & 20 & 20 & -30 & 90  \\
2 & 5 & 10 & 10 & 5	 & 100 & 50  & 30 & 0  &  90 & 80  \\
3 & 5 & 10 & 5  & 5	 & 150 & 80  & 50 & 50 &  90 & 120 \\ [1ex]
\hline
\end{tabular}
\label{table:motherShipControlParamTable}
\end{table}

\newpage
\begin{breakablealgorithm}
  \caption{Mothership Star selection strategy}\label{alg:msstrategyFn}
  \begin{algorithmic}[1]
    \Function{MothershipSelectionStrategy}{$t_{departure}, t, \Delta V$}
    \State
    
    \For{\texttt{j = 1 to 3}} \Comment{For jth mother ship}
    	\State $t_{dep} \gets t_{departure, j}$ 
        \State $t_{arrival} \gets t_{dep} + 2.5 Myr$
        \State $tof \gets t_{arrival} - t_{dep}$
        \State $r_0, v_{0, guess} \gets$ position \& velocity of sol at $t_{dep}$
        \State 
        
 		\State $totalImpulse \gets$ 0
 		\State $impulseCnt \gets$ 0
        \State $numOfViolations \gets $ 0
        \State
        
        
        \While{\var{Constraints are met}}
            \State $R_{arrival},  V_{arrival}\gets$ position \& velocity of all the stars at $t_{arrival}$
            \State

            \State $idx \gets $ \Call{ClosestMomentumStar()}{} 
            \If{\texttt{idx != 0}}
                \State $r_{t}, v_{t} \gets$ position \& velocity of target (idx) at $t_{arrival}$
                \State $[r, v_{0}, v_{f}] \gets$ \Call{ShootingSolver}{$r_0, r_t, tof, v_{0, guess}$}
                \State

                \State $\Delta V_{1} \gets  v_{0} - v_{0, guess} $ \Comment{transfer $\Delta V$}
                \State $\Delta V_{2} \gets v_{t} - v_{f}$  \Comment{rendezvous $\Delta V$}
            \EndIf
            \State
            
            \State \leftcomment{Req. MS or Settlement impulses exceed their limits or No star found}
            \If{ $|\Delta V_{1}| > 200 km/s \OR |\Delta V_{2}| > 300 km/s \OR \texttt{idx == 0}$}
                \State
                \State $tof \gets tof$ + 2.5 Myr
                \State $t_{arrival} \gets t_{dep}$ + tof
                \State
                
                \State $numOfViolations \gets numOfViolations + 1$
                
                \If {$numOfViolations$ > 20}
                	\State \textbf{break}
                \EndIf
                
           \Else \Comment{Successful transfer}
           		\State $\Delta V_{used} \gets \Delta V_{used} + |\Delta V_1| + |\Delta V_2|$
                \State $\Delta V_{max} \gets \Delta V_{max} + 300 km/s$
                \State
                
				\State $totalImpulse  \gets totalImpulse + |\Delta V_1|$
                \State 
                
                \If {$totalImpulse \texttt{ > 500 km/s} \OR t_{arrival} \texttt{+1.5 > 89.5 Myr} $} 
                		\State \textbf{break}
                \Else
                		\State $Stars_1 \gets Stars_1 \cup idx$ \Comment{Add star to Settlement tree}
                \EndIf
                \State 
                
                \State $t_{margin} \gets t_{impulseCnt, j}  - t_{arrival}$ \Comment{time left for next burn}
                \State $t_{dep} \gets t_{arrival} + \max(1.5, t_{margin})$
    			\State 
                
                \State Update states
           \EndIf
        \EndWhile
      \EndFor
    \EndFunction
  \end{algorithmic}
\end{breakablealgorithm}

\begin{figure}[h!]
\includegraphics[width=1\linewidth, height=7cm]{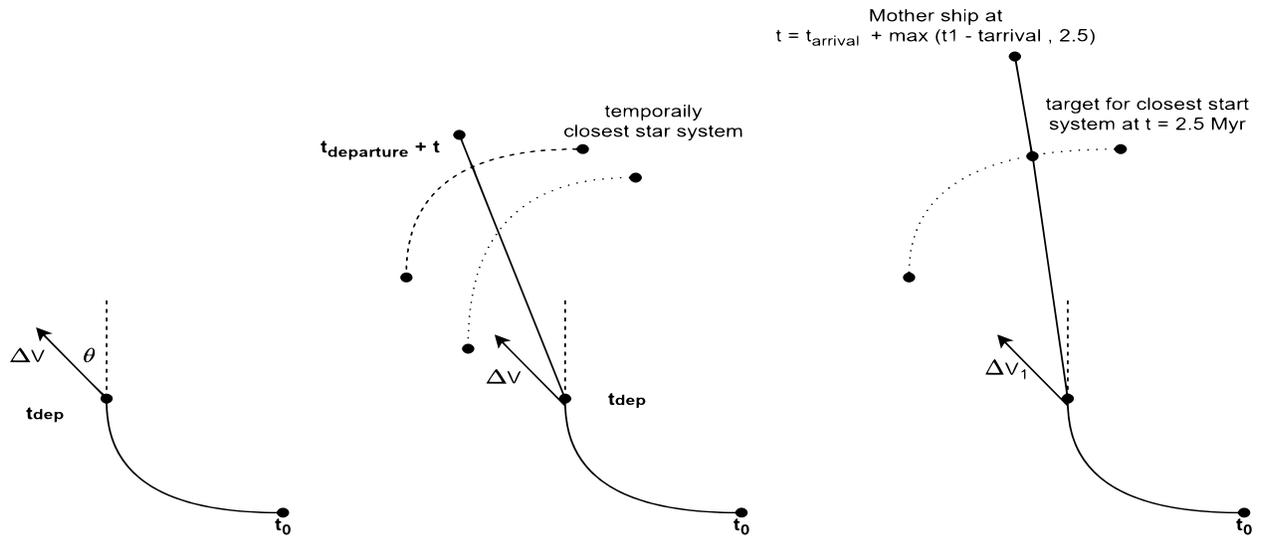} 
\caption{Mother strategy to choose target stars}
\label{fig:msStarSelectionStrategyFig}
\end{figure}

The control parameter $t_{departure}$ decides Myrs after which the mother ship is going to depart from Sol targetting up to three stars after each impulse in its trajectory. Algorithm \ref{alg:msstrategyFn} describes how we capture stars which further get added to the settlement tree, from which settler ships construct their own settlement subtree. To decide the target star system, we start with initial time of flight of 2.5 Myrs. For this time of flight, we capture the trajectory of all the stars. With given $\Delta V$ \& $\Delta \theta$ of mother ship, we find out the set of temporally closest stars within some limits. If this set is empty, then we increase the time of flight by 0.5 Myrs and then we try again. For a non-empty set, we get a successful transfer. A \texttt{ShootingSolver} is used to obtain the exact $\Delta V$ for mother ship and $\Delta V_{rendezvous}$ for settler ships based on the transfer time for this transfer. The target star is added to settlement tree from which settler ships will construct their own settlement subtree. Further, the coasting time after previous impulse determines how much time do we have for next impulse. The successive impulses are performed in similar fashion varying only on control parameters. If the trajectory of the Mother ships are traced based on the control parameters, we would obtained fig. From which we can say that using mother ship we are trying to achieve as much angular distribution as possible. 
\newpage

\section{Settlership strategy}
\vspace{1em}
The settlerships are forced to depart 2 MYr after settling a star. Each star has 3 settler ships which search for the closest relative momentum stars to settle to. The path between settled star and target star is broken into two equal segments of time of flight separated by a mid flight $\Delta V$. An optimization problem called MINDELTAVsolver is used to minimize overall flight time of flight. The $\Delta V_1$,$\Delta V_2$ and $\Delta V_3$ are identified which minimize transfer time. 

\begin{algorithm}[h!]
  \caption{Settler Ship Star selection strategy}\label{alg:ssStrategyFn}
  \begin{algorithmic}[1]
    \Function{SettlershipStrategy}{$id_{fastship}, id_{mothership}$}
    \State
    
    \State ${Stars_{1}} \gets id_{fastship},id_{mothership}$ \Comment{ Define the first generation as the set of stars settled}
    \State $generation \gets 1$ 
    \State 
    
    \While{$generation<20$}
        \State $Stars_{gen}=Stars_{gen}-\{Sol\}$ \Comment{Target stars from the current generation except sol}
        \State
         
        \For{\texttt{j = 1 to $length(Stars_{gen})$}} \Comment{For each settled star}
            \State $t_{delay} \gets 2.5 Myr$
            \State $t_{departure}=t_{arrival}(Stars_{gen_j})+t_{delay}$
            \State 
            
            \State $r_0, v_{0} \gets$ initial position and velocity of $Stars_{gen}$ at $t_{departure}$
            \State
            
            \For{\texttt{k = 1 to 3}} \Comment{For each settled star, 3 settler ships depart to nearby stars}
                \While{}
                    \State $idx \gets$ ClosestMomentumStar($r_0, v_{0},Stars_{gen_{k=1:20}}$) \Comment{Provide initial state and settled star list}
                    \State $t_{arrival} \gets t_{departure} + tof_{guess}$
                    \State $r_{t}, v_{t} \gets$ position \& velocity of target (idx) at $t_{arrival}$
                    \State $t_{burn},\Delta V_1,\Delta V_2,\Delta V_3 \gets$ \Call{MINTIMEsolver}{$r_0,v_0,r_t,v_t,tof_{guess}$}
                    \State
                    
                    \If{$|\Delta V_1| > 150 km/s$ \OR $|\Delta V_2| > 150 km/s$ \OR $|\Delta V_3| > 150 km/s$ \OR $\sum_{n=1}^{3}|\Delta V_n| > 400 km/s$}
                        \State $tof_{guess}=tof_{guess}+1$
                    \Else
                        \State Add $idx$ to $Stars_{gen+1}$
                        \State Update $t_{arrival}(Stars_{gen+1})$
                    \EndIf
                \EndWhile
            \EndFor
        
        \EndFor
    \EndWhile
    \EndFunction
  \end{algorithmic}
\end{algorithm}

\subsection{Settler ship flight time minimization}
\vspace{1em}

The minimization of the overall flight time is performed by the MINTIMEsolver. It requires state of the settled star, time of flight guess and state of the target star as input arguments. Firing constraints of individual $\Delta V$s and cumulative $\Delta V$ for the settler ships are set as the non linear constraints.

\begin{figure}[ht]
\centering
\includegraphics[width=0.6\linewidth]{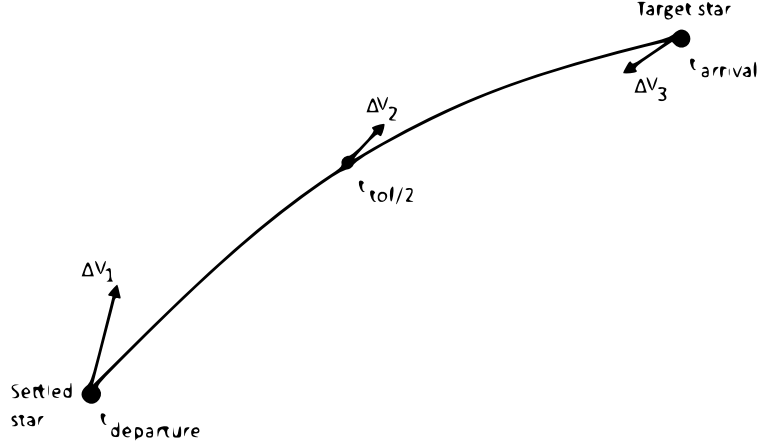} 
\caption{Settler star ships strategy to choose target stars}
\label{fig:settlerShipStrategyFig}
\end{figure}

\newpage

\begin{algorithm}[h!]
  \caption{MINTIMEsolver}\label{alg:solveFn}
  \begin{algorithmic}[1]
    \Function{MINTIMEsolver}{$r_0,v_0,r_t,v_t,tof_{guess}$}
      
        \State $r_0, v_0,r_{mid},v_{mid} \gets$ initial and mid-flight position and velocity of settler ship
        \State Define State $x=[r_{init},v_{init},r_{mid},v_{mid},tof_{first_segment},tof_{second_segment}]$ \Comment{States contain the first and the next segment initial states and flight duration}
        \State $t_{arrival} \gets t_{departure} + 2.5 Myr$
        \While{$\Delta V$ constraints are met}
            \State minimize (x(13)+x(14)) \Comment{Total time of flight}
            \State where,
                \State $r_{init} \gets r_0$
                \State $r_{final} \gets r_f$ \Comment{Propagated end positions from Delta Vs}
                \State $\Delta V_1<175 km/s$ and $\Delta V_2<175 km/s$ and $\Delta V_3<175 km/s$ and $\sum_{n=1}^{3}\Delta V_n<400 km/s$ 
            \State $idx \gets $ ClosestMomentumStar()
        \EndWhile
      \State \textbf{return}
    \EndFunction
  \end{algorithmic}
\end{algorithm}

\subsection{Maximum expansion strategy}
\vspace{1em}

The maximum expansion strategy is covered by the star selection algorithm referred under ClosestMomentumStar(). The strategy involves looking in the direction of travel of the currently settled star and search for an unsettled star which is closest to the momentum (cross product of position and velocity) of the currently settled star. The search is done by alternatively increasing the radius and oscillating the angle w.r.t to the current direction.

\section{Results}
\vspace{1em}

This section combines the mother ship, Fast ship and Settler ship strategies into one. The three mother ships depart Sol at 0,5 and 5 MYr respectively. The delta V direction from the departure Sol velocity direction is kept as 20 deg, 0 deg and 50 deg respectively to separate their trajectories spatially as much as possible, each trajectory bent in the clockwise direction due to the galactic spiral motion direction. The mother ships settle three stars each and the settler ship strategy is allowed to proceed from there on. The fast ships are sent two adjacent quadrants near the edge of the galaxy. A fast expansion happens along the edge of the galaxy. Notice that the top-left areas which remain largely deserted since our strategy constrains the mother ships not to go there and there is not enough time to expand in particularly that direction. Another important observation is that since settler ships are targeting high value stars the expansion frontiers are merge and that stops a lot of potential growth from the settled stars which cannot release settler ships to nearby areas. However, the expansion process over 13 generations reaches to significantly spread regions of the galaxy.

\begin{figure}[H]
    \centering
    \begin{subfigure}[b]{0.33\linewidth}
        \includegraphics[width=\linewidth]{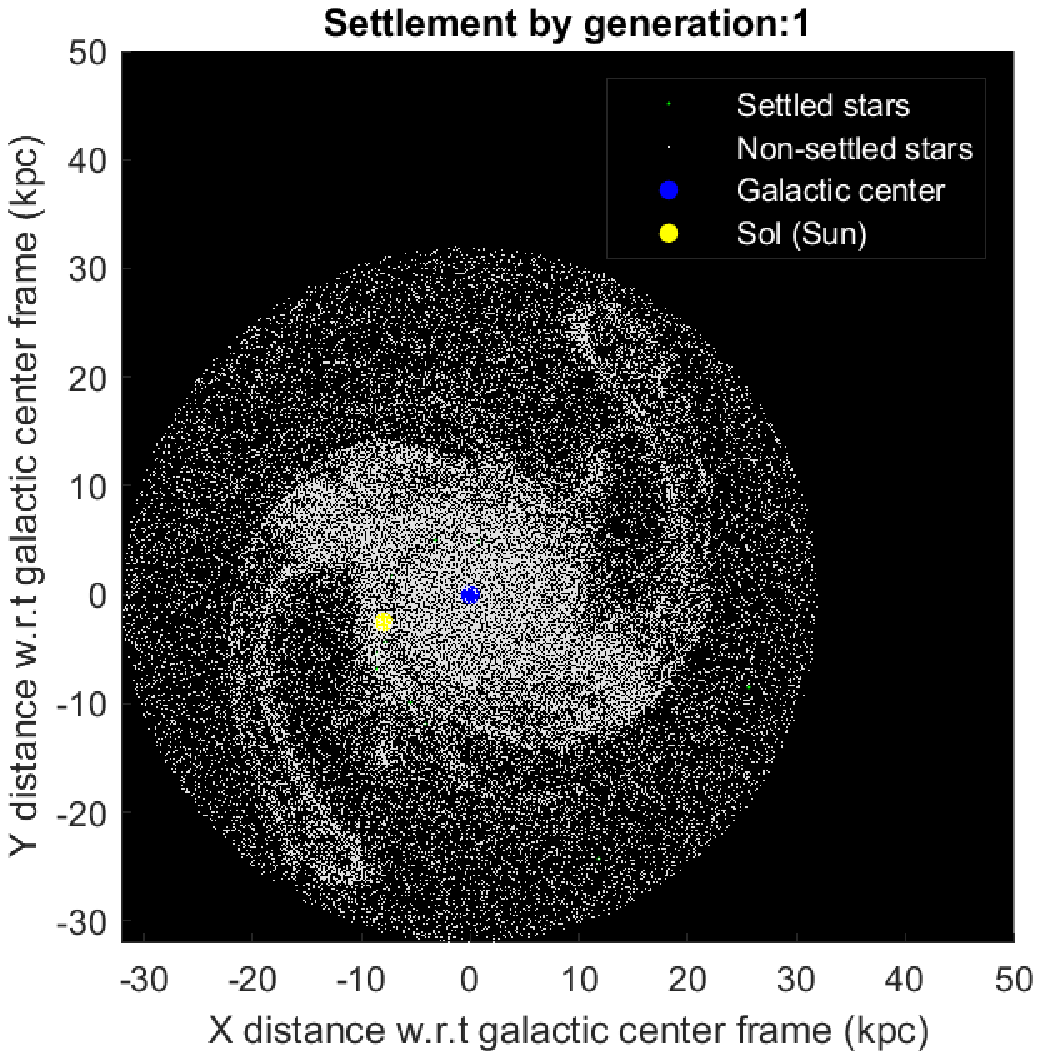}
        \caption{1st Generation}
    \end{subfigure}
    \begin{subfigure}[b]{0.33\linewidth}
        \includegraphics[width=\linewidth]{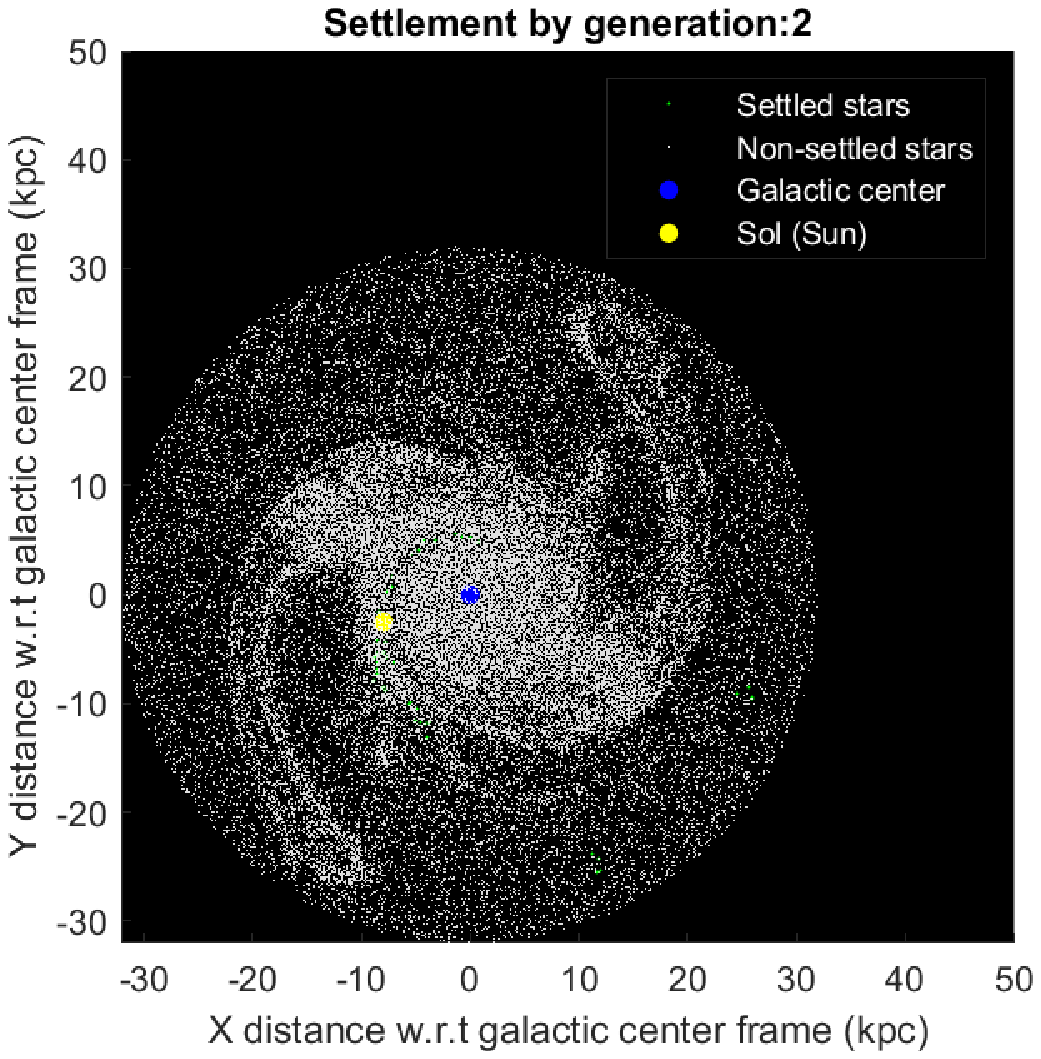}
        \caption{2nd Generation}
    \end{subfigure}
    \begin{subfigure}[b]{0.33\linewidth}
        \includegraphics[width=\linewidth]{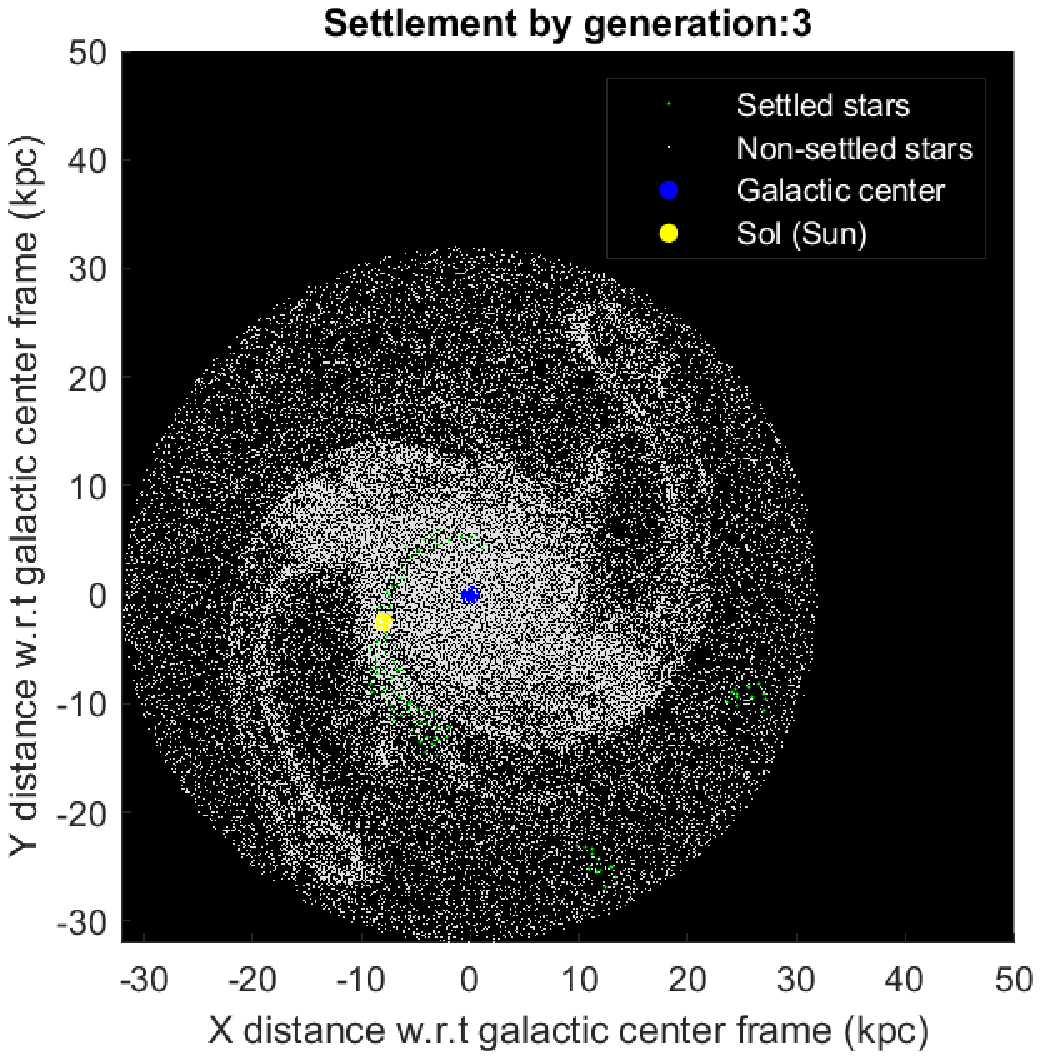}
        \caption{3rd Generation}
    \end{subfigure}
    \begin{subfigure}[b]{0.33\linewidth}
        \includegraphics[width=\linewidth]{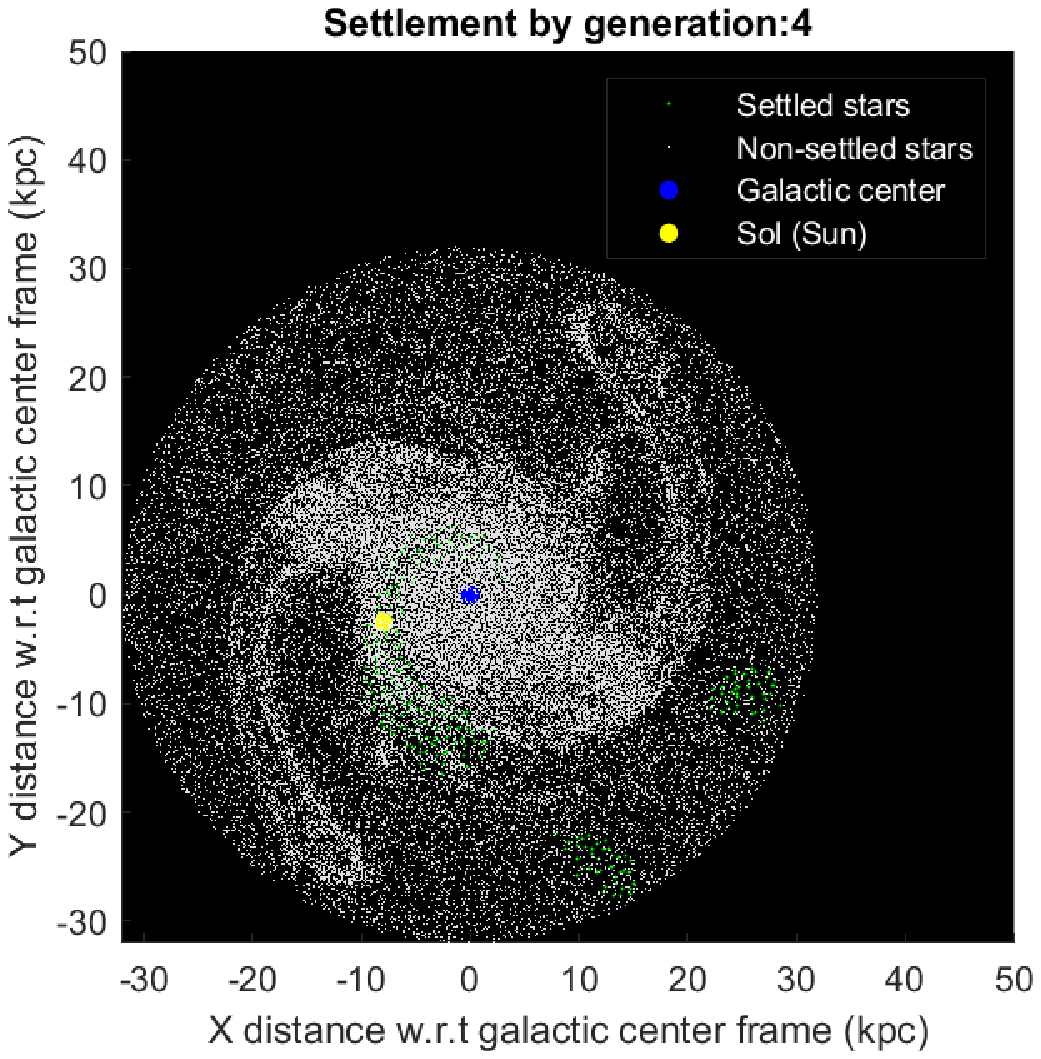}
        \caption{4th Generation}
    \end{subfigure}
    \begin{subfigure}[b]{0.33\linewidth}
        \includegraphics[width=\linewidth]{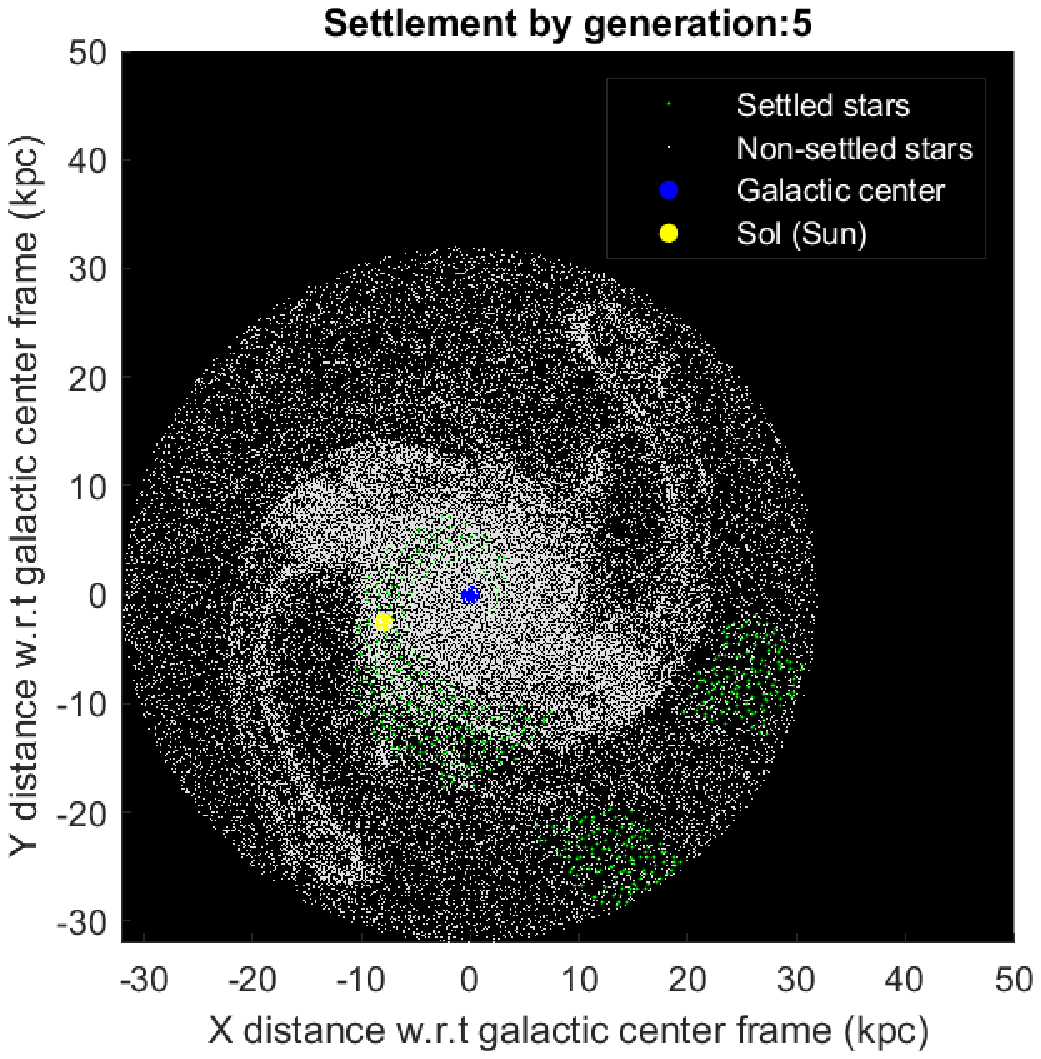}
        \caption{5th Generation}
    \end{subfigure}
    \begin{subfigure}[b]{0.33\linewidth}
        \includegraphics[width=\linewidth]{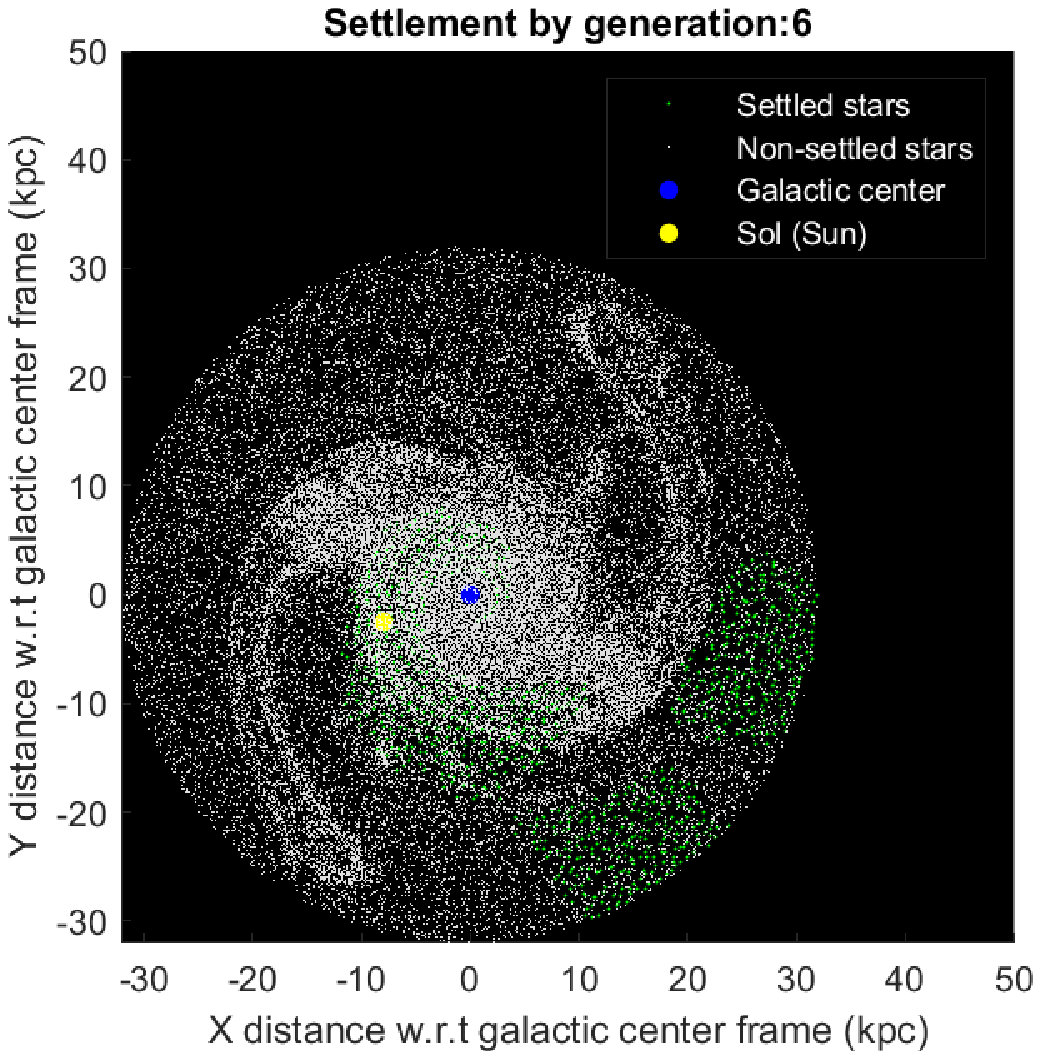}
        \caption{6th Generation}
    \end{subfigure}
    \begin{subfigure}[b]{0.33\linewidth}
        \includegraphics[width=\linewidth]{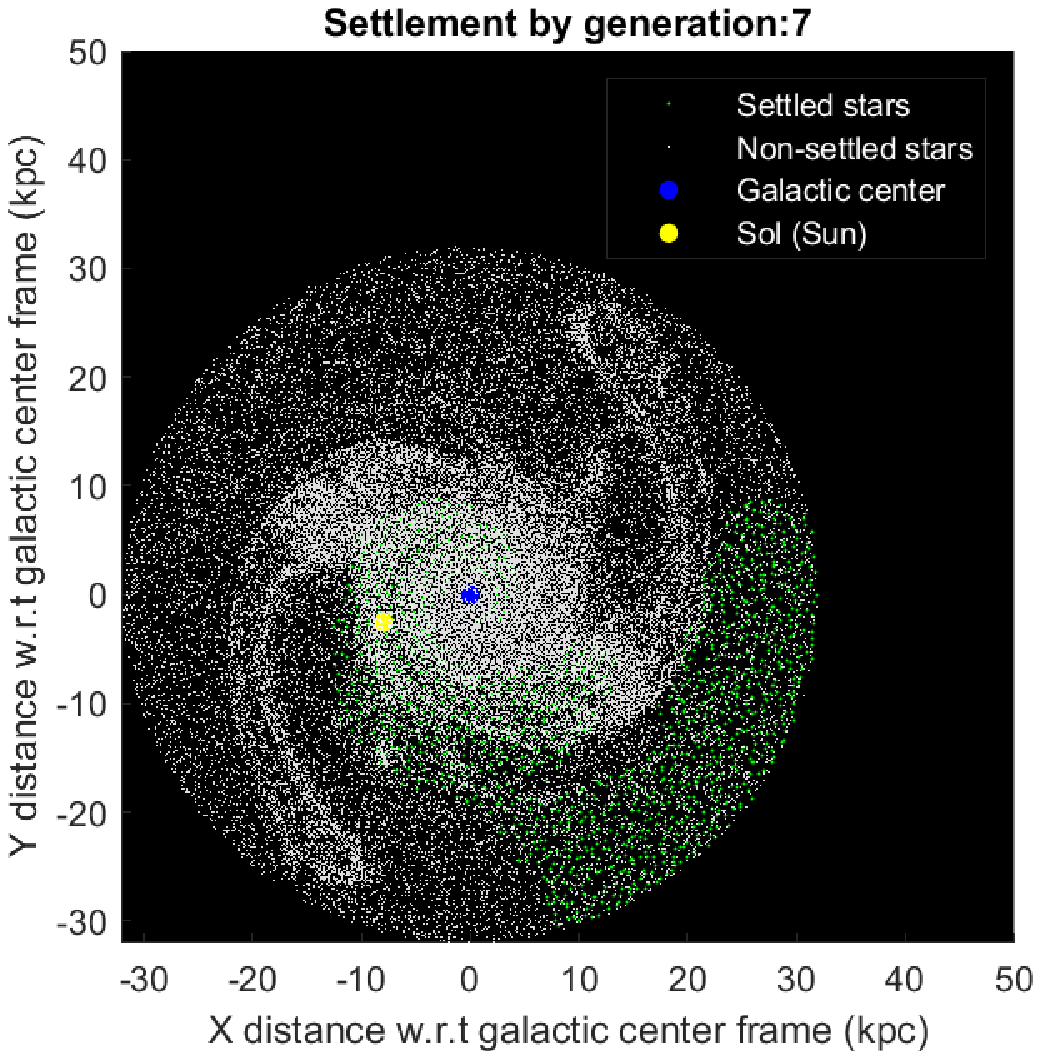}
        \caption{7th Generation}
    \end{subfigure}
    \begin{subfigure}[b]{0.33\linewidth}
        \includegraphics[width=\linewidth]{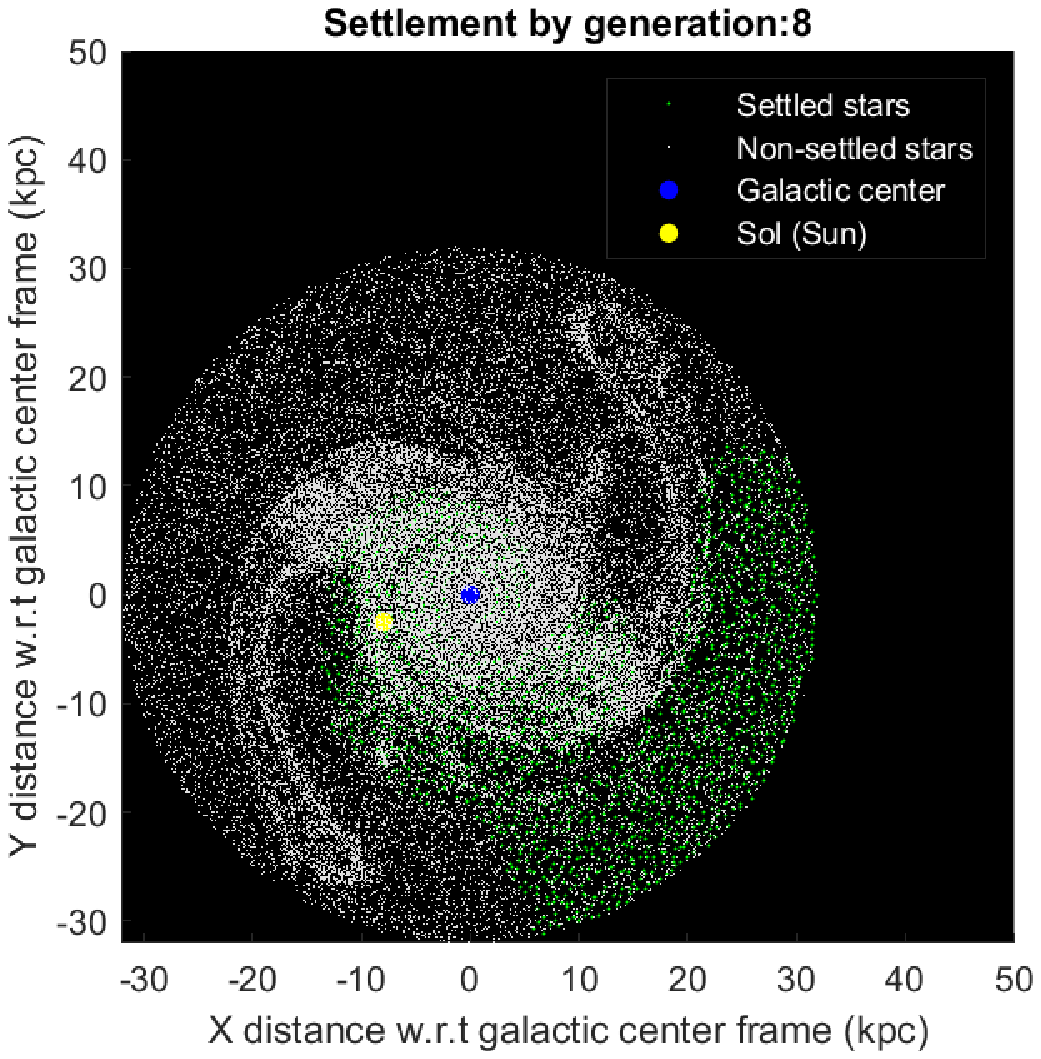}
        \caption{8th Generation}
    \end{subfigure}
    \begin{subfigure}[b]{0.33\linewidth}
        \includegraphics[width=\linewidth]{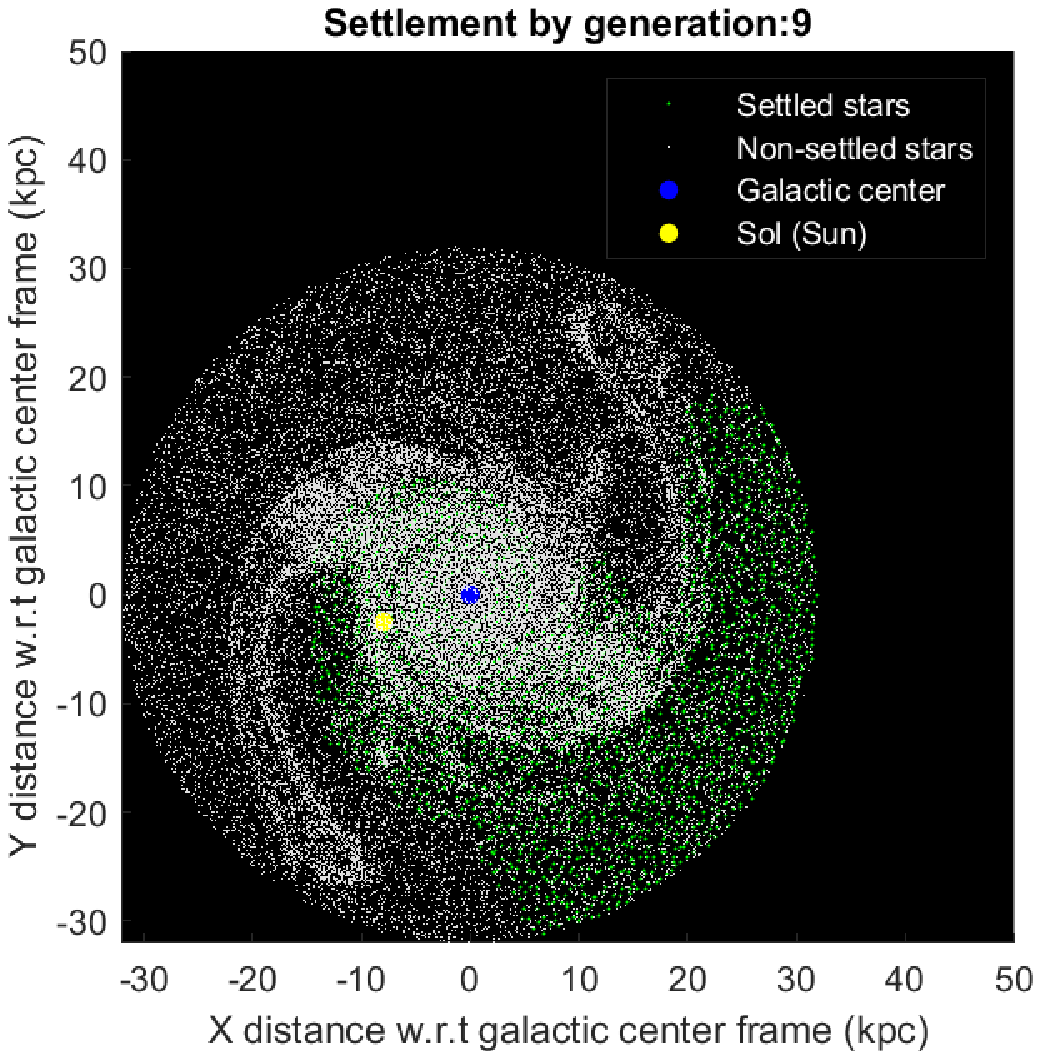}
        \caption{9th Generation}
    \end{subfigure}
    \begin{subfigure}[b]{0.33\linewidth}
        \includegraphics[width=\linewidth]{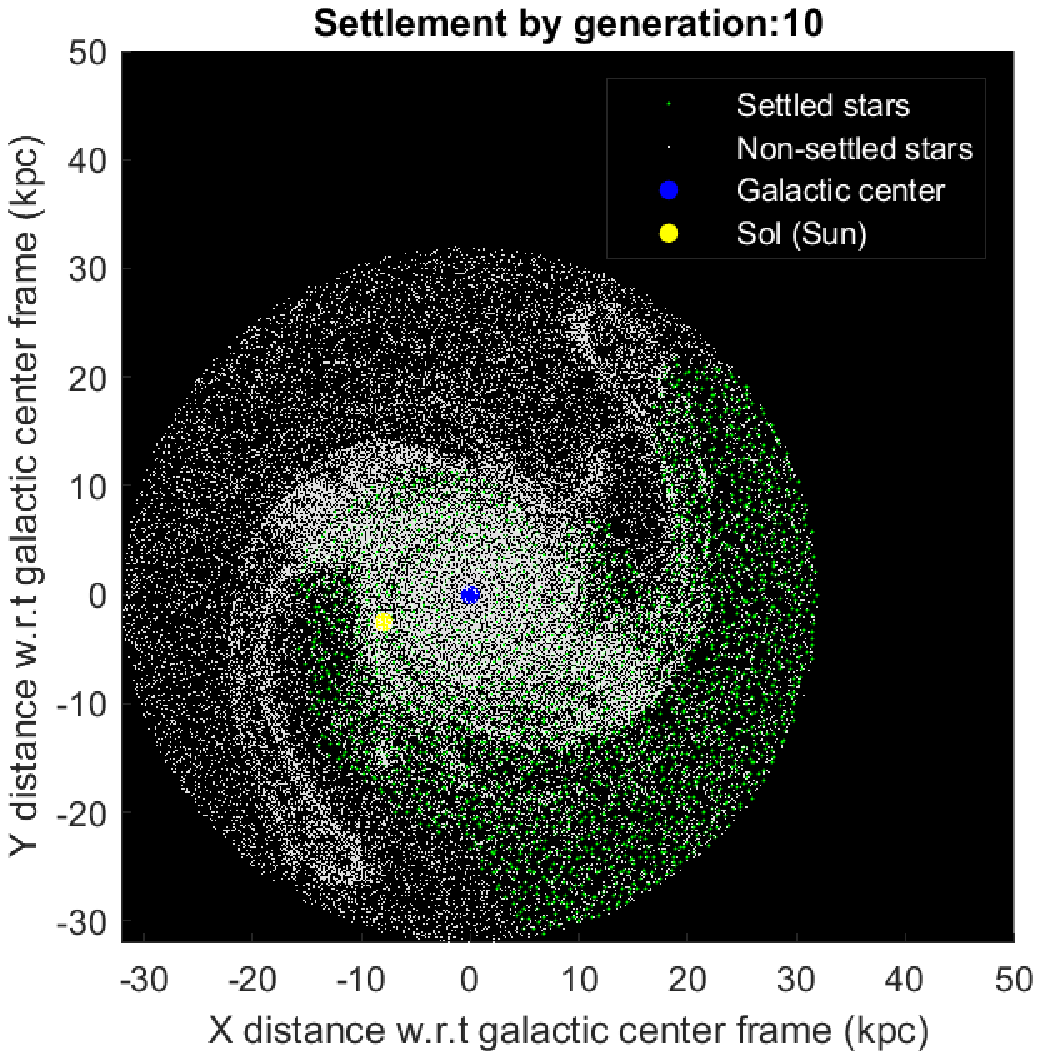}
        \caption{10th Generation}
    \end{subfigure}
    \begin{subfigure}[b]{0.33\linewidth}
        \includegraphics[width=\linewidth]{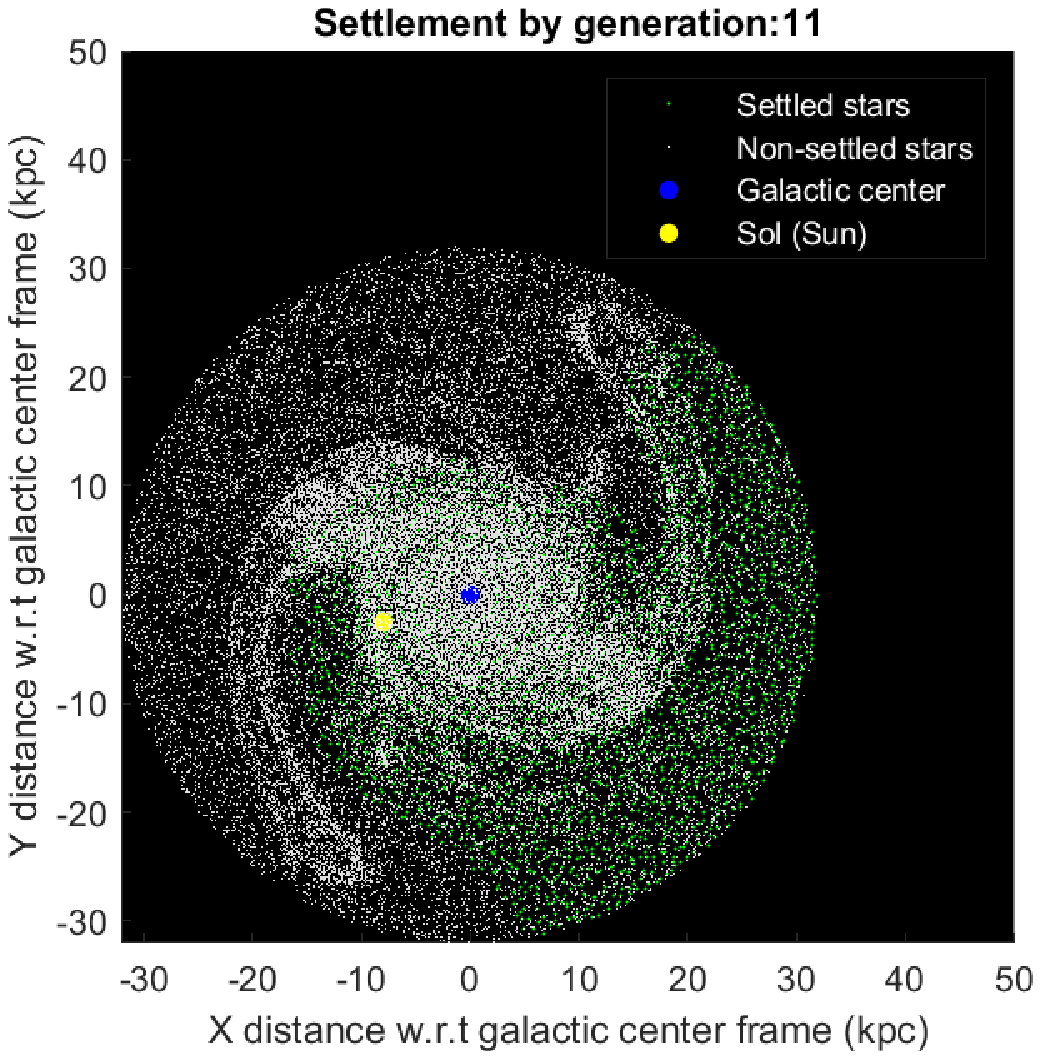}
        \caption{11th Generation}
    \end{subfigure}
    \begin{subfigure}[b]{0.33\linewidth}
        \includegraphics[width=\linewidth]{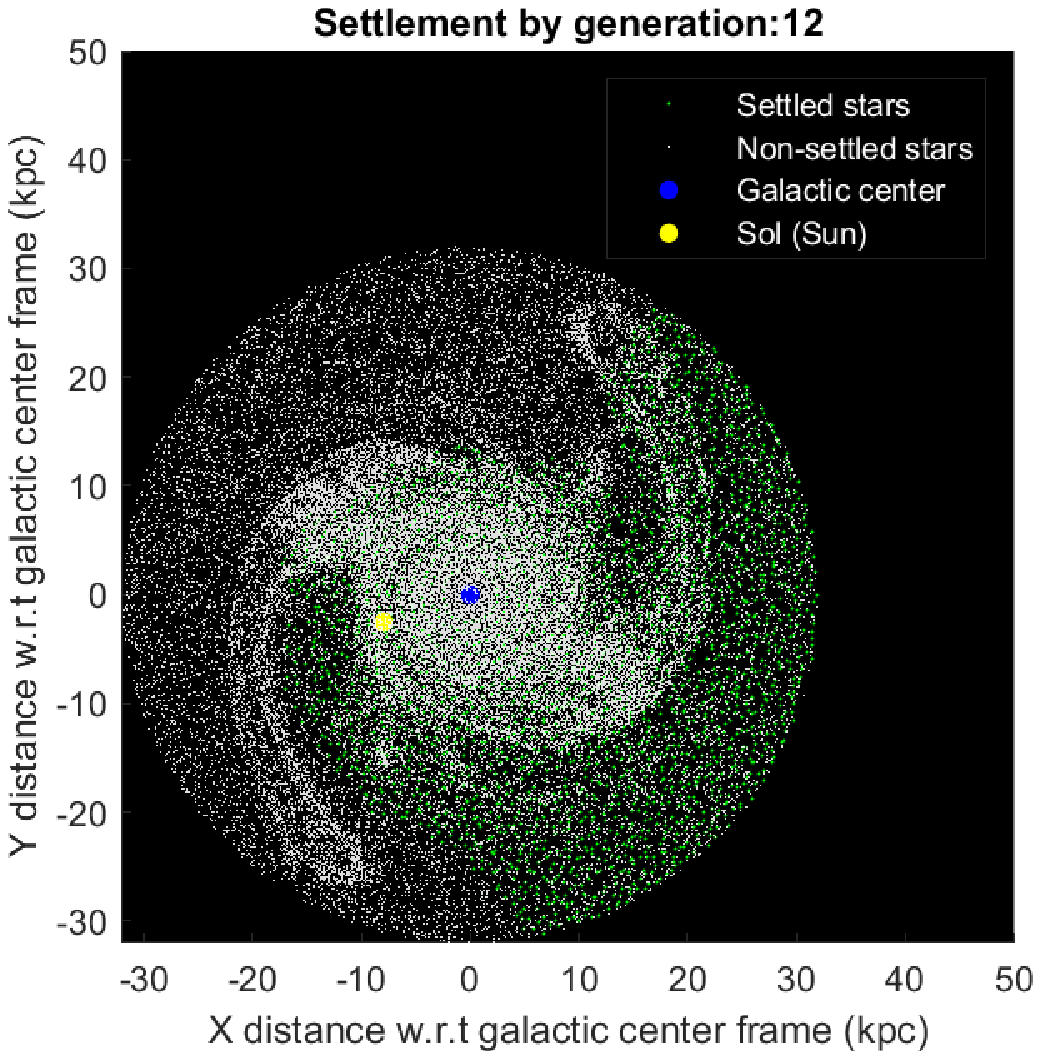}
        \caption{12th Generation}
    \end{subfigure}
    \caption{Star settlement with generation}
    \label{fig:settlementWithGenFig}
\end{figure}

\begin{figure}[h!]
    \centering
    \includegraphics[width=0.6\linewidth]{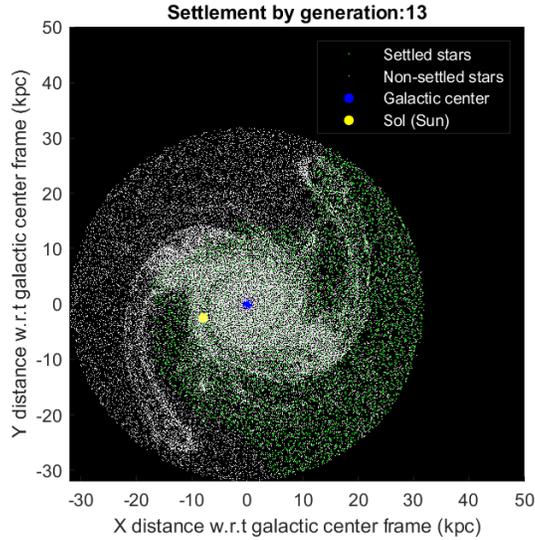}
    \caption{13 generations of settlement as seen at $t=t_0 + 90$ Myr}
    \label{fig:13GenerationsFig}
\end{figure}

\begin{figure}[hbt!]
    \centering
    \includegraphics[width=0.6\linewidth]{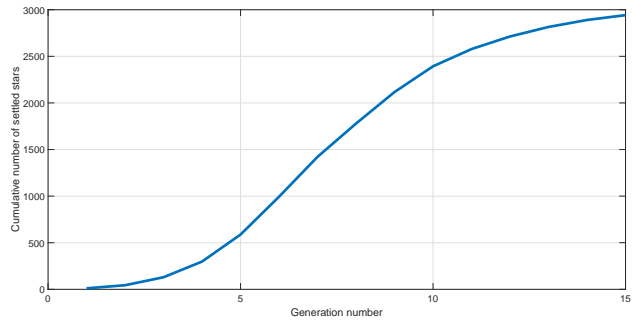}
    \caption{Cumulative star settled vs Generation}
    \label{fig:cumulativeStarsVsGenFig}
\end{figure}

\section{Conclusion}
\vspace{1em}

The paper summarizes the strategy adopted by our team- Karmarkar's Gang during the GTOC X competition. The strategy involves a spatially separated set of mothership trajectories settling three stars each, fast ships populating the edges of the galaxy for a uniform radial distribution. The expansion of the settlement is implemented by generations of settler ships who target close high-value stars and gradually expand to nearly 3000 stars. The merit function $J$ reaches to about 150.

We find that two major areas of improvement would be in the targeting of multiple stars by each mothership and target mothership trajectories to target a more even distribution w.r.t polar angle. 

\section{Code}
All the associated code has been pushed in a Github repository \cite{Karmakar19}.

\bibliography{sample}

\end{document}